\documentclass[final,5p,times,twocolumn]{elsarticle}

\usepackage{amssymb}
\usepackage{amsmath}
\usepackage{multirow}
\usepackage{color} 
\usepackage{wrapfig}
\usepackage{graphicx}
\usepackage{flushend}

\journal{Neurocomputing}

\begin{document}

\begin{frontmatter}



\title{Structure Modeling Activation Free Fourier Network for Spacecraft Image Denoising}


\author[a]{Jingfan Yang} 
\author[a]{Hu Gao} 
\author[a]{Ying Zhang} 
\author[a]{Bowen Ma}
\author[a]{Depeng Dang\corref{cor1}} 

\affiliation[a]{organization={School of Artificial Intelligence, Beijing Normal University},
            addressline={No.19, Xinjiekouwai St.}, 
            city={Beijing},
            postcode={100875}, 
            country={China}}

\cortext[cor1]{Corresponding author: Depeng Dang (ddepeng@bnu.edu.cn)}

\begin{abstract}
Spacecraft image denoising is a crucial fundamental technology closely related to aerospace research. However, the existing deep learning-based image denoising methods are primarily designed for natural image and fail to adequately consider the characteristics of spacecraft image(e.g. low-light conditions, repetitive periodic structures), resulting in suboptimal performance in the spacecraft image denoising task. To address the aforementioned problems, we propose a Structure modeling Activation Free Fourier Network (SAFFN), which is an efficient spacecraft image denoising method including Structure Modeling Block (SMB) and Activation Free Fourier Block (AFFB). We present SMB to effectively extract edge information and model the structure for better identification of spacecraft components from dark regions in spacecraft noise image. We present AFFB and utilize an improved Fast Fourier block to extract repetitive periodic features and long-range information in noisy spacecraft image. Extensive experimental results demonstrate that our SAFFN performs competitively compared to the state-of-the-art methods on spacecraft noise image datasets. The codes are available at: https://github.com/shenduke/SAFFN. 
\end{abstract}




\end{frontmatter}



\section{Introduction}
In recent years, space imaging technology has undergone remarkable advancements, enabling an increasingly crucial role for spacecraft image in diverse aerospace-related researches, such as spacecraft pose estimation~\cite{liu2022neural, zhang2024monocular}, spacecraft 3D reconstruction~\cite{wang2015simulation, chang2024reconstructing}, spacecraft control~\cite{sharma2020neural, dong2022learning} and spacecraft debris reentry prediction~\cite{kyselica2023towards}. In order to better complete the aforementioned aerospace researches, the high-quality spacecraft image is essential. However, due to the harsh conditions of the outer space environment, spacecraft image is often degraded by noise~\cite{murray2021mask}, leading to the performance degradation of various space missions. Hence, it is necessary to research the denoising of spacecraft image.

So far, some spacecraft image denoising approaches have been proposed, such as Adaptive Noise Template Prediction (ANTP)-based denoising~\cite{miao2022image} and Dual-Cycle GAN-based denoising~\cite{chen2022single}. Meanwhile, there are a large number of deep learning-based general image denoising methods~\cite{ zhang2021accurate, xu2024stacked} and radar image denoising methods~\cite{liu2020sar, pan2024sar}. But these methods lack designs according to the characteristics of spacecraft noise image.

Through analysis, we found that the spacecraft noise image has the following characteristics. Firstly, spacecraft is in a harsh space lighting environment, and the surface illumination of the spacecraft changes constantly~\cite{cao2023detection}, resulting in many obtained spacecraft images appear to present low-light. Secondly, spacecraft is the human-made object, so the overall structure of spacecraft is relatively regular, and there are a lot of repetitive periodic components such as the solar panel. Finally, spacecraft image application scenarios, such as spacecraft control, require high real-time performance.

\begin{figure}[t]
\centering
\includegraphics[width=1.0\linewidth]{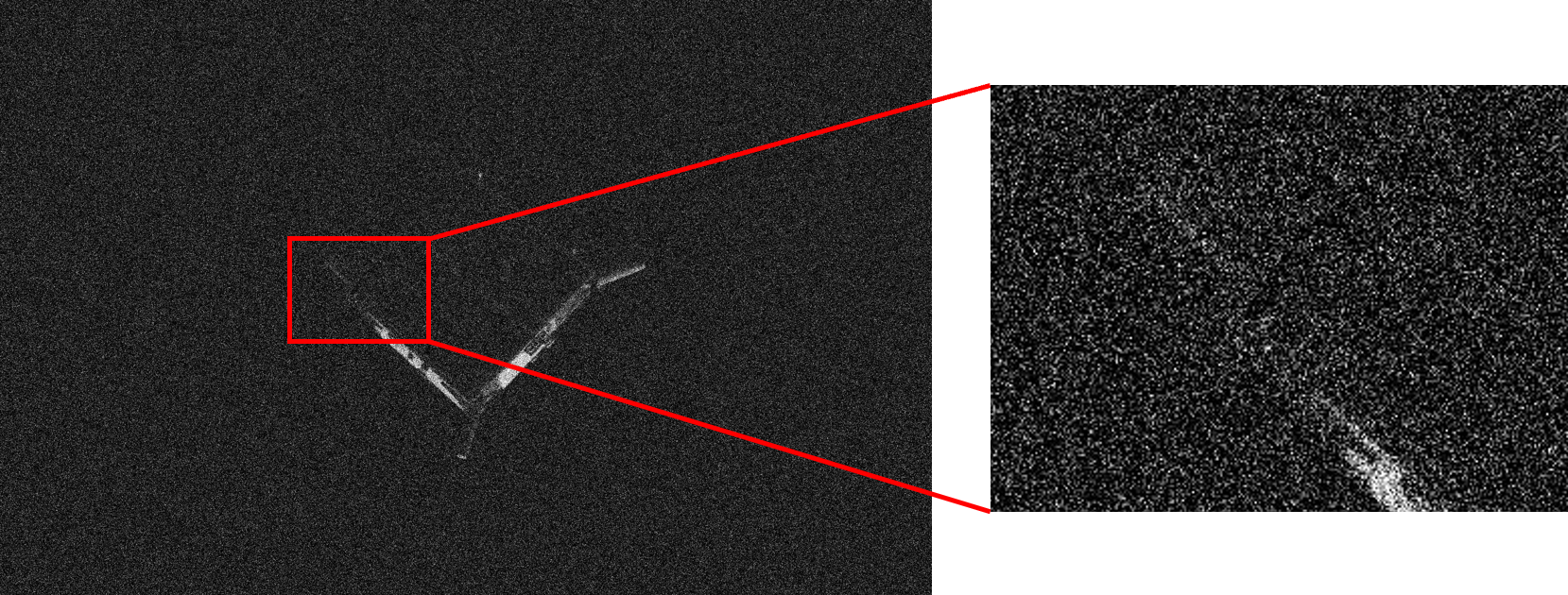}
\caption{Example of spacecraft noise image.}
\label{fig:example}
\end{figure}

In low-light spacecraft noise images, as illustrated in Fig. \ref{fig:example}, it is challenging to distinguish the spacecraft components in the dark regions from the surrounding background, which adversely affects the performance of spacecraft image denoising. Some studies~\cite{rana2021edge, zhu2020eemefn} suggest leveraging structural information, particularly edge features, to enhance denoising performance through distinguishing different parts of these dark regions. Among the commonly used edge extraction operators (Prewitt, Sobel, Canny, and Laplacian), the Sobel operator is distinguished by its strong anti-noise ability and high computational efficiency~\cite{guo2018feature}. However, the conventional Sobel operator is fixed-valued which limits its adaptability and training capability. So we need to design a trainable edge convolution based on the Sobel operator.

As human-made object, there are a large number of repetitive periodic structures in spacecraft, and the Fast Fourier Transform is very suitable for capturing these periodic structures because the harmonics within the spectrograms are essentially periodic~\cite{sun2023lightweight}. Furthermore, Fast Fourier Transform can extract  long-range global information. Therefore, we will design a novel and efficient Fast Fourier convolution block in our model.

In this paper, we propose Structure modeling Activation Free Fourier Network (SAFFN), which is a novel and efficient method spacecraft image denoising method and successfully resolves the above issues in spacecraft image denoising. Inspired by~\cite{chen2022simple}, we design an efficient nonlinear activation free network as the foundation of our SAFFN. Then, we present Structure Modeling Block (SMB), which can effectively extract the structural information of low-light spacecraft image by the trainable Edge Convolution. Additionally, we present Activation Free Fourier Block (AFFB) to capture the periodic features and long-range information in spacecraft image by Simplified Fast Fourier Block (SFFB).

Overall, the contributions of this paper are summarized below:
\begin{itemize}
    \item We propose a novel Structure modeling Activation Free Fourier Network (SAFFN) for effective spacecraft image denoising. SAFFN is a highly efficient method and better adapts to the characteristics of spacecraft image denoising.
    \item We present Structure Modeling Block (SMB) to effectively extract the edge features and model the structure, significantly improving the ability to distinguish spacecraft components from the background in noisy spacecraft image.
    \item We present Activation Free Fourier Block (AFFB) to capture the repetitive periodic features and long-range global information in spacecraft image.
\end{itemize}

This paper is organized as follows. Section \ref{rel_work} introduces the related works. Section \ref{method} presents our proposed method. Section \ref{exp} provides the experimental results and related analysis. Section \ref{conclusion} reports the conclusion. 

\begin{figure*}[t]
\centering
\includegraphics[width=1.0\linewidth]{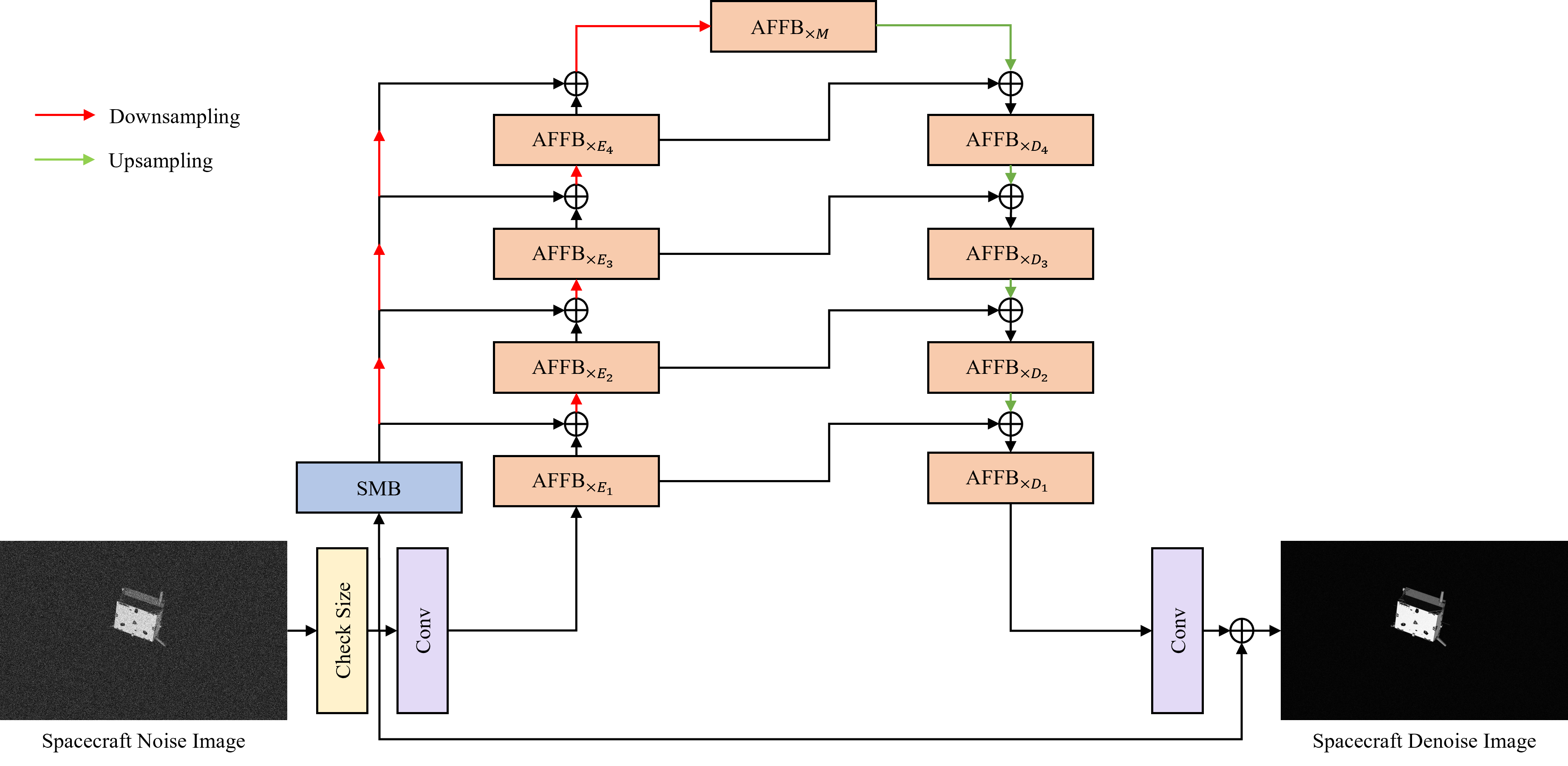}
\caption{The overall architecture of Structure modeling Activation Free Fourier Network(SAFFN).}
\label{fig:saffn}
\end{figure*}

\section{Related works}
\label{rel_work}
\subsection{Spacecraft image restoration}
Spacecraft image restoration tasks aim to restore a degraded spacecraft image to a clean one. Zhang et al.~\cite{zhang2019comparable} comparably study four spacecraft image SR models based on SRCNN, FSRCNN, VDSR and DRCN, and fine-tune the SR model using space objects images. Feng et al.~\cite{feng2019super} propose a deep CNN method based on optimization to solve spacecraft image SR problem, and use Nesterov Accelerated Gradient (NAG) to accelerate the convergence of loss function to avoid the large oscillation of stochastic gradient descent. SRDN~\cite{yang2021srdn} is an end-to-end spacecraft image SR and deblurring network based on ESRGAN and RaGAN, which uses contrastive learning to extract degradation feature and symmetrical downsampling and upsampling modules to restore the texture information. Zhou et al.~\cite{zhou2022self} propose a self-supervised method for spacecraft video SR, and use a pyramid structure to train from coarse to fine. Miao et al.~\cite{miao2022image} introduce adaptive noise template prediction and gray analysis of local area to achieve spacecraft image denoising. Dual-Cycle GAN~\cite{chen2022single} is a spacecraft image denoising and SR method, which use unpaired images to improve the quality of spacecraft images. Shi et al.~\cite{shi2023super} propose a spacecraft image SR method based on dual regression and deformable convolutional attention mechanism to better extract useful features in space object images. Yang et al.~\cite{yang2023ssrn} propose an space SR network (SSRN), which utilize the non-local sparse attention to capture the long-range self-similarity in the spacecraft ISAR image and design the residual atrous spatial pyramid pooling block to extract the multi-scale features. Gao et al.~\cite{gao2024learning} propose a sparse and selective feature fusion network for spacecraft ISAR image SR, and design the trainable top-k selection operator that retains the most critical attention scores to enhance the distinction from the background. However, these methods overlook the impact of low-light conditions and repetitive structures on spacecraft image denoising.

\subsection{Image denoising}
Image denoising aims to reduce the noise from a variety of sources in images. With the rapid development of deep learning, a large number of deep learning based image denoising methods have been proposed~\cite{  jia2019focnet, xu2022hyperspectral}. These methods can be classified into three categories: CNN-based method~\cite{zhang2017beyond, sun2023multi}, attention-based method~\cite{shi2021hyperspectral, zhou2023deep} and Transformer-based method~\cite{ gao2024prompt, jiang2024edformer}.

CNN was first introduced into image denoising. Xu et al.~\cite{xu2015patch} propose the first CNN-based image denoising method which combines stacked autoencoder and dropout together to reduce the impact of overfitting. Jifara et al.~\cite{jifara2019medical} introduce residual learning and design a deep feed forward CNN for image denoising. Jia et al.~\cite{jia2021ddunet} present a novel image denoising method DDUNet which connects the feature maps in each level cross cascading U-Nets to relieve the vanishing gradient problem and encourage feature reuse. Zhang et al. ~\cite{zhang2021plug} propose DRUNet which applies skip connection between the upsampling and downsampling layers at each scales of U-Net to handle various noise levels. MWDCNN~\cite{tian2023multi} is a multi-level image denoising CNN based on wavelet transform, and uses a dynamic convolution for making a tradeoff between denoising performance and computational costs. Xu et al.~\cite{xu2024pan} propose Pan-Denoising network, which leverages panchromatic image to guide HSI denoising and introduces the Panchromatic Weighted RCTV (PWRCTV) regularization to enhance the noise suppression. 

Tian et al.~\cite{tian2020attention} firstly introduce attention mechanism into image denoising, and design an attention-guided denoising CNN named ADNet. Zhang et al.~\cite{zhang2021accurate} propose using global channel attention to maintain the main scaling channel information to guide adversarial training. Mei et al.~\cite{mei2023pyramid} present a novel Pyramid Attention module which captures long-range features from a multi-scale pyramid to borrow clean signals from the coarser levels. Wu et al.~\cite{wu2024dual} propose a novel Dual-branch Residual Attention Network (DRANet) for image denoising, which extracts rich local features between different convolution layers through two different parallel residual attention branches. Liu et al.~\cite{liu2024lg} introduce a novel SAR image denoising approach LG-DBNet, which utilizes a dual-branch structure—one branch for local feature extraction through hybrid attention mechanism, and another for global feature extraction with self-attention.

Due to the strong global information extraction capability of Transformer, Transformer-based methods have also been applied to image denoising. Chen et al.~\cite{chen2021pre} develop the first Transformer-based model for image denoising. Liang et al.~\cite{liang2021swinir} propose an image denoising method (SwinIR) based on Swin Transformer, which uses the shifted window mechanism to achieve long-range dependencies modeling. Zamir et al.~\cite{zamir2022restormer} propose an efficient Transformer-based image denoising model Restormer which can aggregate the local and non-local pixel interactions through multi-Dconv head transposed attention module. SERT~\cite{li2023spectral} is a hyperspectral image denoising rectangle Transformer, and uses horizontal and vertical rectangle self-attention to extract self-similarity in image spatial domain. Li et al.~\cite{li2024ewt} propose the Efficient Wavelet-Transformer (EWT), which integrates Discrete Wavelet Transform (DWT) with the Transformer architectures to enhance the computational efficiency while preserving denoising performance. Tian et al.~\cite{tian2024heterogeneous} propose the Heterogeneous Window Transformer (HWformer), which introduces heterogeneous global windows to capture richer contextual information while maintaining computational efficiency. By incorporating the window shift mechanism and the sparse feed-forward network, HWformer enhances both local and global feature extraction, improving denoising performance. Choi et al.~\cite{choi2024reciprocal} propose a lightweight Reciprocal Attention Mixing Transformer (RAMiT), which utilizes the dimensional reciprocal attention mixing Transformer Transformer (D-RAMiT) module to parallelly compute different multi-head self-attention and hierarchical reciprocal attention mixing (H-RAMi) layer to maintain the efficient hierarchical structure.

\section{Methodology}
\label{method}

Let $I$ represent a clean spacecraft image with $H \times W$ pixels. When the image is degraded by additive noise $\mathcal{N}$ from the harsh outer space environment, such as the Gaussian noise. The observed noise image $I_{noisy}$ can be mathematically modeled as:
\begin{eqnarray}
    I_{noisy} = I + \mathcal{N}
\end{eqnarray}%
Our primary objective is to create an efficient spacecraft image denoising model to learn the mapping from $I_{noisy}$ to $I$. 

Considering this goal, we propose SAFFN which includes Structure Modeling Block (SMB) and Activation Free Fourier Block (AFFB) to achieve effective spacecraft image denoising by modeling structural information and periodic repetition features. First, we present the overall architecture of our SAFFN. Subsequently, we give the detailed descriptions of SMB and AFFB. The proposed AFFB comprises Simplified Fast Fourier Block (SFFB) and Simple Gate.

\begin{figure*}[t]
\centering
\includegraphics[width=0.8\linewidth]{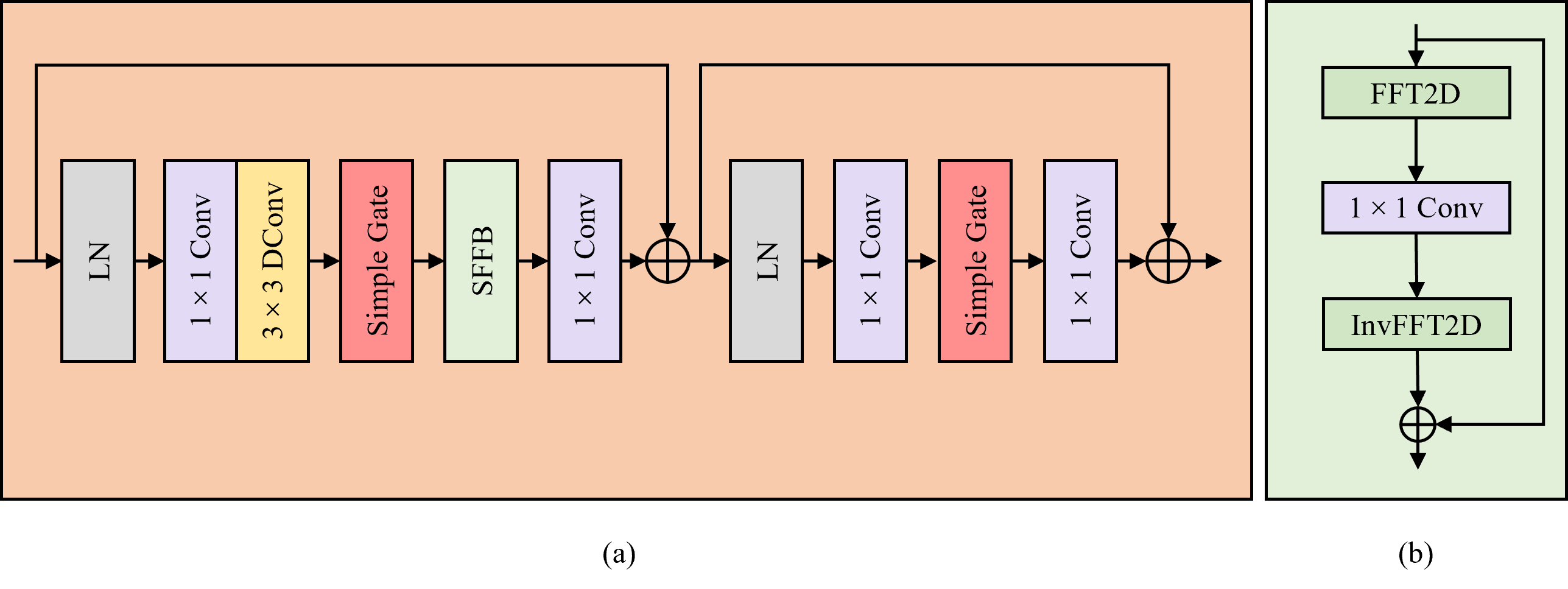}
\caption{(a)The architecture of Activation Free Fourier Block (AFFB). (b)The architecture of Simplified Fast Fourier Block (SFFB).}
\label{fig:affb}
\end{figure*}

\subsection{Overall architecture}
 The overall architecture of SAFFN is shown in Fig. \ref{fig:saffn}. For the input spacecraft noise image $I_{noisy}$, we first utilize a $3\times3$ convolution layer to extract the shallow feature maps, and an SMB to extract the structural features. Then, these shallow feature maps are processed by the 5-level encoder-decoder to obtain deep features. The encoder-decoder consists of several AFFBs. In the encoder, the input of each downsampling layer includes not only the previous AFFB output features but also the SMB output structure features with corresponding scale. We also add skip connections to ensure the stability of training. Finally, we apply a $3\times3$ convolution layer to deep features to generate the residual image, which is added to noisy spacecraft image to obtain denoising spacecraft images.
 We optimize our SAFFN using PSNR loss:
\begin{eqnarray}
    PSNR~Loss = - 10 \cdot log_{10}\frac{(MAX^2)}{||I_d-I_{gt}||^2+\epsilon}
\end{eqnarray}%
where $MAX$ is the maximum possible pixel value of the image, $I_d$ denotes the obtained denoising spacecraft image, $I_{gt}$ denotes the ground-truth spacecraft image, and $\epsilon$ is the infinitesimal to avoid the denominator being 0.

\subsection{Structure Modeling Block (SMB)}
There are a large number of low-light images in the spacecraft image dataset. Many spacecraft components are difficult to distinguish from the background due to the low-lighting conditions. In order to solve this problem, we propose the SMB, which accurately models the reliable structural information of spacecraft components in low-light areas by effectively extracting the edge features. Formally, the process of our SMB is represented as: 
\begin{eqnarray}
    X_{SMB} = DEConv(EConv(I_{noisy}))
\end{eqnarray}%
where $I_{noisy}$ is the input noisy spacecraft image, $X_{SMB}$ is the output structure feature, $EConv($$\cdot$$)$ and $DEConv($$\cdot$$)$ respectively denote $3\times3$ Edge Convolution and $3\times3$ Depthwise Edge Convolution. 

\begin{figure}[t]
\centering
\includegraphics[width=0.6\linewidth]{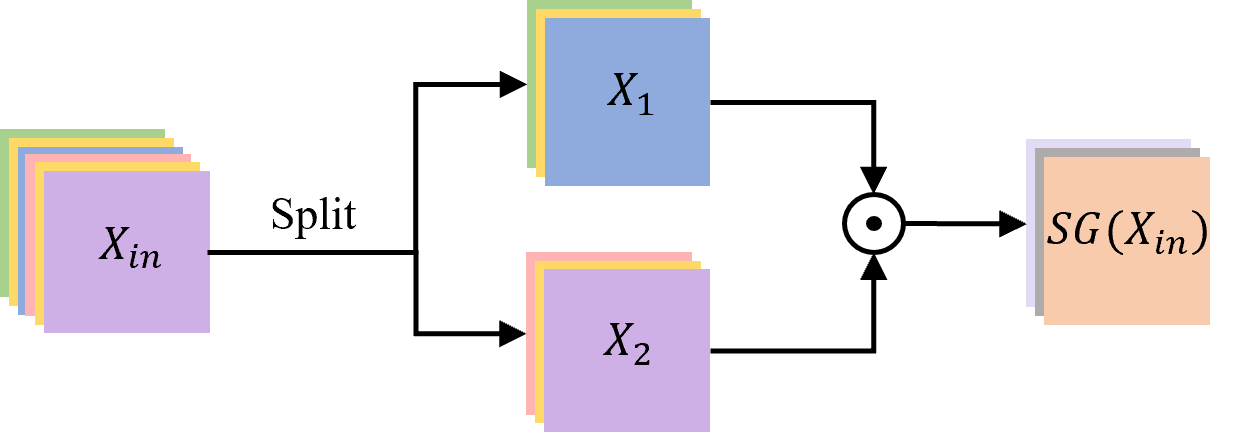}
\caption{Illustration of Simple Gate.}
\label{fig:sg}
\end{figure}

\begin{table*}[!ht]
\caption{Quantitative results on SPEED dataset. The best and the second best results are marked in \textcolor{red}{red} and \textcolor{blue}{blue}.}
\label{tab:speed_results}
\centering
\begin{tabular}{lllllllll}
\hline
\multirow{2}{*}{Methods}& \multicolumn{2}{c}{$\sigma$=50}   & \multicolumn{2}{c}{$\sigma$=75} & \multicolumn{2}{c}{$\sigma$=100} & \multirow{2}{*}{MACs} & \multirow{2}{*}{Runtime} \\ 
  & \multicolumn{1}{c}{PSNR$\uparrow$}  & \multicolumn{1}{c}{SSIM$\uparrow$} & \multicolumn{1}{c}{PSNR$\uparrow$}  & \multicolumn{1}{c}{SSIM$\uparrow$} & \multicolumn{1}{c}{PSNR$\uparrow$}  & \multicolumn{1}{c}{SSIM$\uparrow$} & \\
\hline
DnCNN~\cite{zhang2017beyond} & 28.44 & 0.6548 & 28.06 & 0.5718 & 27.04 & 0.5808 & 36.73G & 5.86ms \\
DRUNet~\cite{zhang2021plug} & 31.79 & 0.6904 & 31.22 & 0.6757 & 30.72 & 0.6647 & 143.69G & 215.32ms\\
MWDCNN~\cite{tian2023multi} & 31.77 & 0.6896 & 31.17 & 0.6736 & 30.68 & 0.6627 & 215.56G & 263.76ms\\
ADNet~\cite{tian2020attention} & 30.41 & 0.6201 & 30.63 & 0.6479 & 29.58 & 0.6254 & 34.30G & 176.52ms\\
DRANet~\cite{wu2024dual} & 31.81 & 0.6906 & 31.21 & 0.6753 & 30.75 & 0.6649 & 148.88G & 299.52ms\\
SwinIR~\cite{liang2021swinir} & \textcolor{blue}{31.83} & \textcolor{blue}{0.6909} & \textcolor{blue}{31.25} & \textcolor{blue}{0.6760} & \textcolor{blue}{30.79} & \textcolor{blue}{0.6657} & 216.25G & 1513.61ms\\
SERT~\cite{li2023spectral} & 31.74 & 0.6880 & 31.16 & 0.6726 & 30.69 & 0.6617 & 103.45G & 535.66ms\\
EWT~\cite{li2024ewt} & 31.76 & 0.6883 & 31.19 & 0.6731 & 30.73 & 0.6627 & 189.78G & 1087.78ms\\
RAMiT~\cite{choi2024reciprocal} & 31.78 & 0.6888 & 31.20 & 0.6743 & 30.69 & 0.6602 & 34.74G & 486.82ms\\
SAFFN(Our approach) & \textcolor{red}{31.86} & \textcolor{red}{0.6920} & \textcolor{red}{31.32} & \textcolor{red}{0.6775} & \textcolor{red}{30.87} & \textcolor{red}{0.6673} & 86.36G & 44.95ms\\
\hline
\end{tabular}
\end{table*}

\subsubsection{Edge convolution}
Sobel operator is an important edge detection method in image processing due to the robustness against noise. However, the conventional sobel operator is fixed-valued and cannot be trained. To address this limitation, we design Edge Convolution inspired by sobel operator. In Edge Convolution, we rewrite the convolution kernel into four kinds of sobel operator forms, and introduce a learnable factor in edge convolution kernel to adaptively extract the edge information of different intensities. 

Specifically, we use the vertical edge convolution kernel $EK_{1}$ to extract the vertical edge features when the output channel index $c \bmod 4=0$.
\begin{eqnarray}
EK_{1} = 
\begin{bmatrix} -\gamma & -2\gamma & -\gamma \\ 
0 & 0 & 0 \\
\gamma & 2\gamma & \gamma\end{bmatrix}
\end{eqnarray}
where $\gamma$ is the learnable factor. 

We use the horizontal edge convolution kernel $EK_{2}$ to extract horizontal edge features when $c \bmod 4=1$.
\begin{eqnarray}
EK_{2} = 
\begin{bmatrix} -\gamma & 0 & \gamma \\ 
-2\gamma & 0 & 2\gamma \\
 -\gamma & 0 & \gamma\end{bmatrix}
\end{eqnarray}

We use the diagonal edge convolution kernel $EK_{3}$ and $EK_{4}$ to extract diagonal edge features in two directions when $c \bmod 4=2$ and $c \bmod 4=3$ respectively.
\begin{eqnarray}
EK_{3} = 
\begin{bmatrix} -2\gamma & -\gamma & 0 \\ 
-\gamma & 0 & \gamma \\
 0 & \gamma & 2\gamma\end{bmatrix}
\end{eqnarray}
\begin{eqnarray}
EK_{4} = 
\begin{bmatrix} 0 & \gamma & 2\gamma \\ 
-\gamma & 0 & \gamma \\
 -2\gamma & -\gamma & 0\end{bmatrix}
\end{eqnarray}

Then, performing convolution operation by using the aforementioned edge convolution kernel can better model the edges and extract structural information of low-light noisy spacecraft image. Formally, the detailed process of our Edge Convolution is represented as:
\begin{eqnarray}
    EC(i,j) = \sum_x\sum_y I_{noisy}(i+x, j+y) EK_n(x,y)
\end{eqnarray}%
where $I_{noisy}$$\in$$\mathbb{R}^{H \times W \times C_{in}} $ is the input image of Edge Convolution, $H$, $W$ and $C_{in}$ respectively represent the height, width and channel of the input image, $EK_n $$\in$$\{EK_1,EK_2,EK_3,EK_4\}$ is the edge kernel selected based on the output channel index, $EC$$\in$$\mathbb{R}^{H \times W \times C_{out}} $ is the output feature map of edge Convolution, $C_{out}$ represent the number of the output feature maps (the output channel number), $i$ and $j$ represent the coordinate on the output feature map $EC$, $x$ and $y$ represent the relative coordinate inside the edge kernel $EK_n$, $i+x$ and $j+y$ represent the coordinate of the corresponding element in the input image $I_{noisy}$.

\begin{figure*}[t]
\centering
\begin{minipage}[t]{0.7\linewidth}
  \centering
  \includegraphics[width=1.0\textwidth]{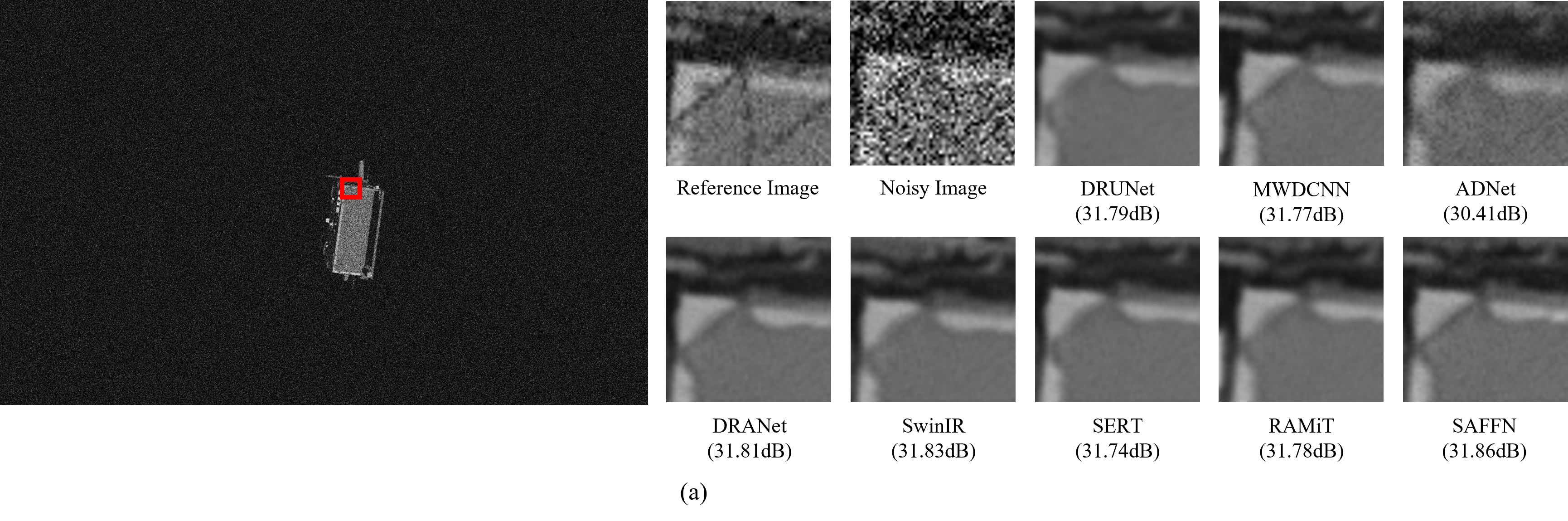}
\end{minipage}  

\begin{minipage}[t]{0.7\linewidth}
  \centering
  \includegraphics[width=1.0\textwidth]{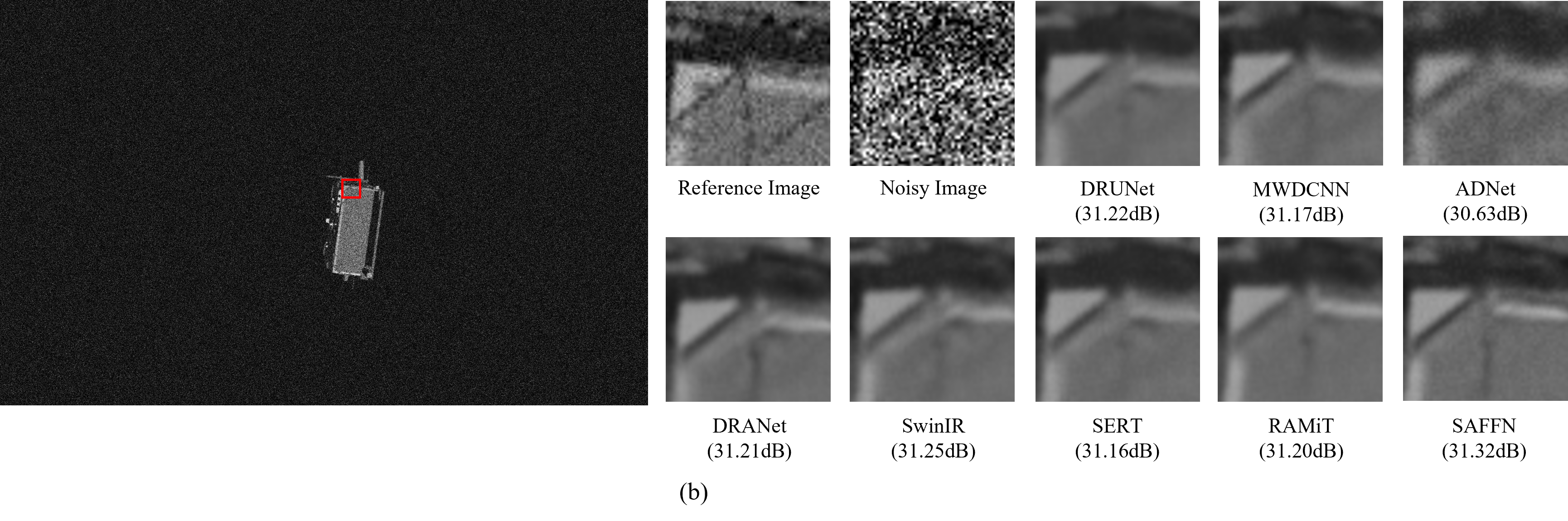}
\end{minipage}  

\begin{minipage}[t]{0.7\linewidth}
  \centering
  \includegraphics[width=1.0\textwidth]{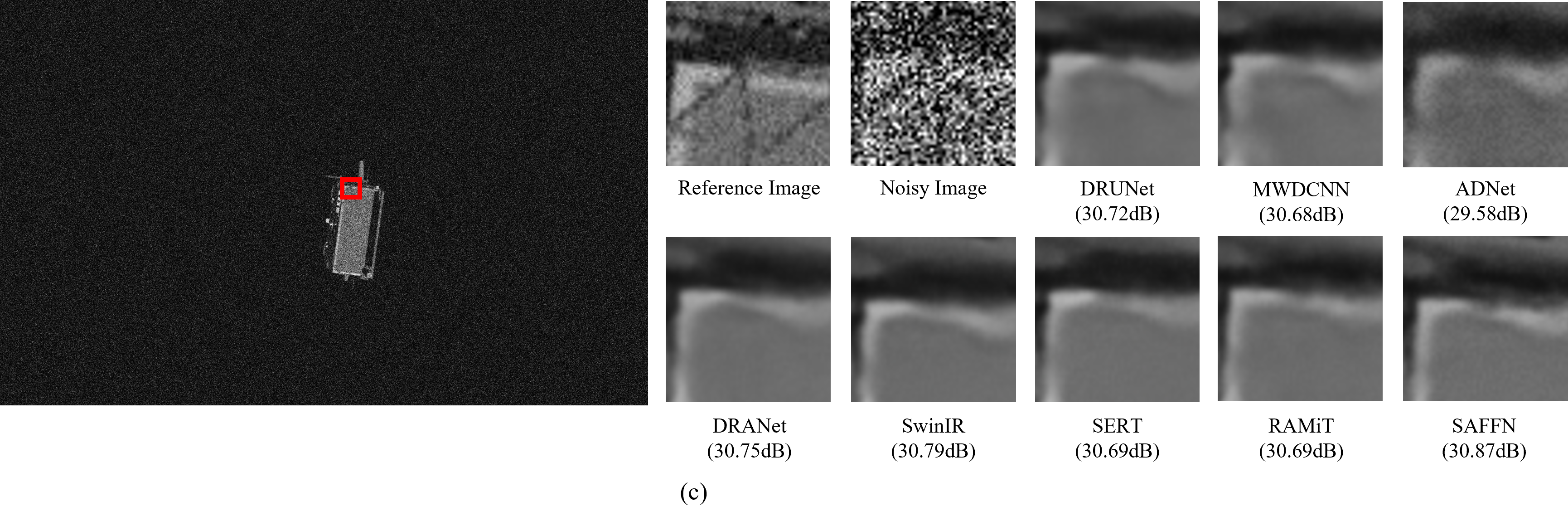}
\end{minipage}  
\caption{Visual comparison on SPEED - 1. (a): $\sigma$ = 50. (b): $\sigma$ = 75. (c): $\sigma$ = 100.}
\label{fig:visual_results_img1}
\end{figure*}

\begin{figure*}[t]
\centering
\begin{minipage}[t]{0.7\linewidth}
  \centering
  \includegraphics[width=1.0\textwidth]{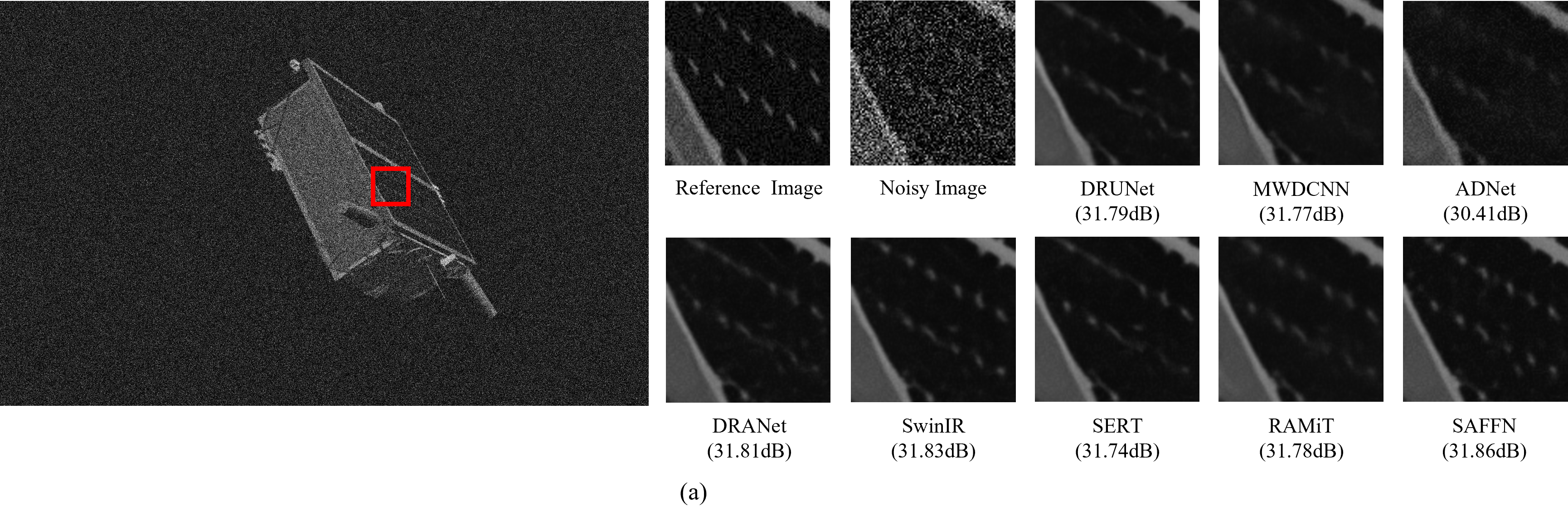}
\end{minipage}  

\begin{minipage}[t]{0.7\linewidth}
  \centering
  \includegraphics[width=1.0\textwidth]{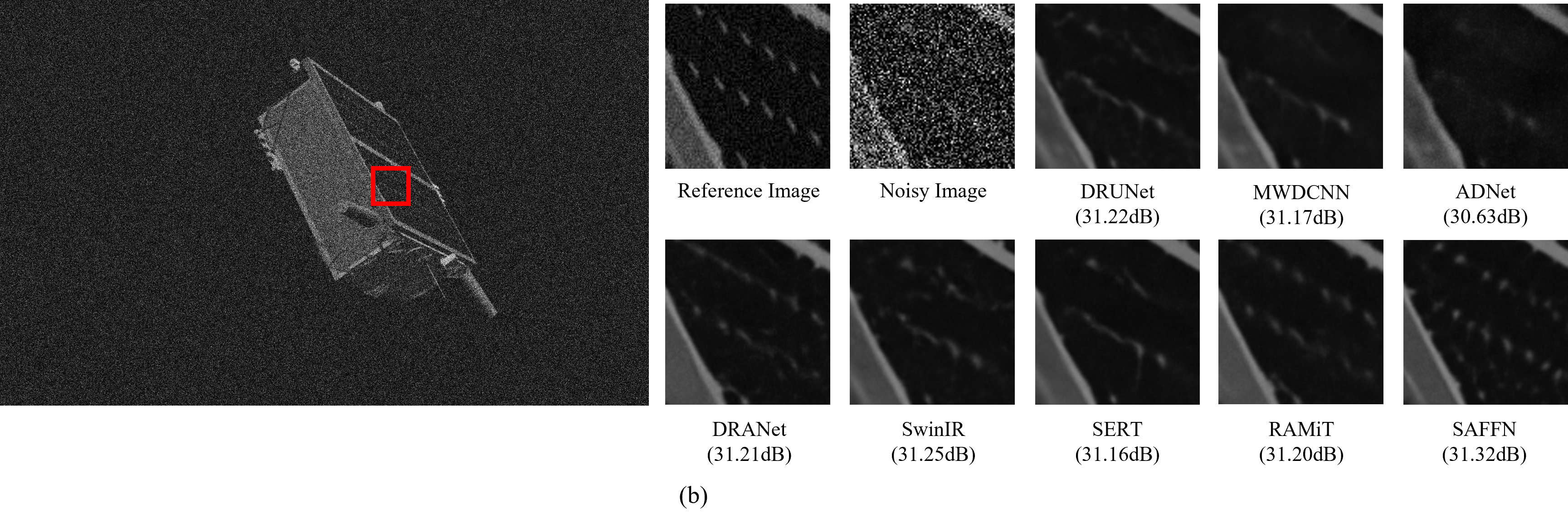}
\end{minipage}  

\begin{minipage}[t]{0.7\linewidth}
  \centering
  \includegraphics[width=1.0\textwidth]{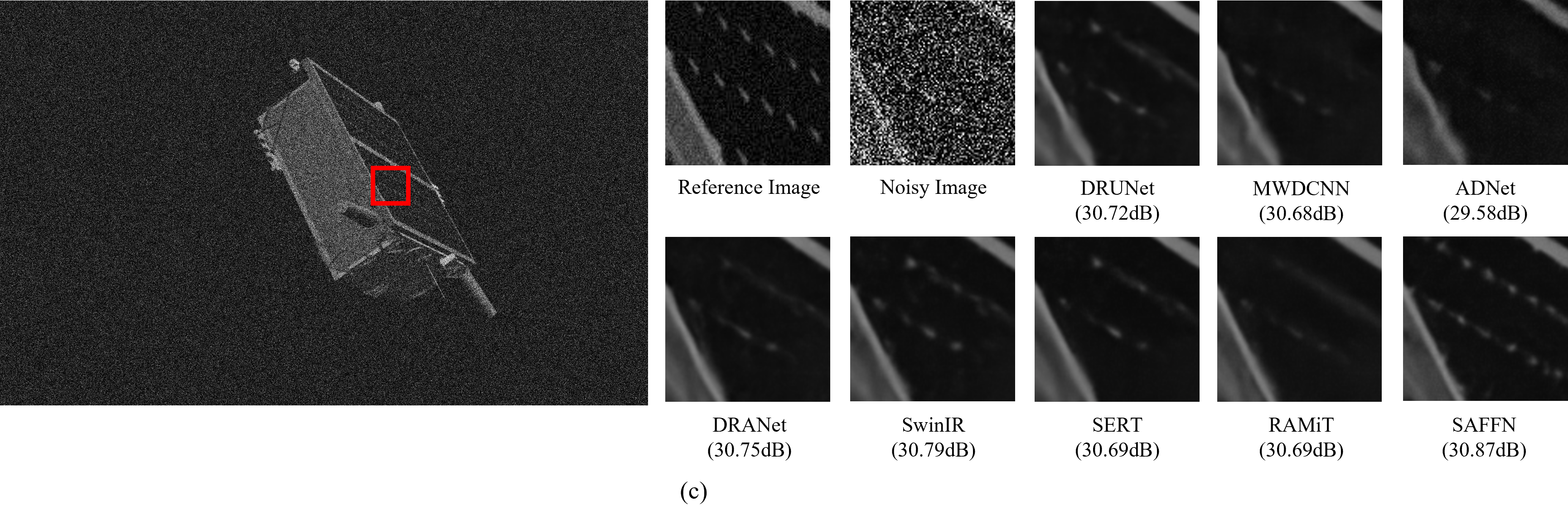}
\end{minipage}  
\caption{Visual comparison on SPEED - 2. (a): $\sigma$ = 50. (b): $\sigma$ = 75. (c): $\sigma$ = 100.}
\label{fig:visual_results_img2}
\end{figure*}

\subsubsection{Depthwise edge convolution}
The composition of the Depthwise Edge Convolution is similar to that of the Edge Convolution. But in the Depthwise Edge Convolution, one edge kernel is responsible for one channel, and one channel is only convolved by one edge kernel.

\subsection{Activation Free Fourier Block (AFFB)}
In order to efficiently extract features especially repetitive periodic structural features from spacecraft noise image, we design the AFFB. In AFFB, we did not use any nonlinear activation functions to reduce the computational complexity. The architecture of AFFB is shown in Fig. \ref{fig:affb} (a).

Our AFFB consists of convolution, Depthwise convolution, Simplified Fast Fourier Block (SFFB) and Simple Gate. Among them, SFFB extracts repetitive periodic features and long-range global features, Simple Gate can replace nonlinear activation functions in performance while reducing the computational complexity. Formally, the detailed process of our AFFB is represented as: 
\begin{eqnarray}
    X_{SG_1} = SG(DConv(Conv(LN(X_{in}))))
\end{eqnarray}
\begin{eqnarray}
    X_{SFFB} = SFFB(X_{SG_1})
\end{eqnarray}
\begin{eqnarray}
    X_{F} = Conv(X_{SFFB})+\alpha*X_{in}
\end{eqnarray}
\begin{eqnarray}
    X_{SG_2} = SG(Conv(LN(X_{F})))
\end{eqnarray}
\begin{eqnarray}
    X_{AFFB} = Conv(X_{SG_2})+\beta*X_{F}
\end{eqnarray}
where $X_{in}$ and $X_{AFFB}$ are respectively the input and output feature maps of AFFB, $LN($$\cdot$$)$ denotes the layer normalization layer, $Conv($$\cdot$$)$ and $DConv($$\cdot$$)$ respectively denote convolution layer and depthwise convolution layer, $SG($$\cdot$$)$ denotes Simple Gate, $SFFB($$\cdot$$)$ denotes SFFB, $\alpha$ and $\beta$ are both trainable parameters that control the weight of skip connection.

\subsubsection{Simplified Fast Fourier Block (SFFB)}
Spacecraft is a human-made object, and have a large number of repetitive periodic structures such as solar panel. Fast Fourier transform is very suitable for capturing these periodic structures. Hence, we propose SFFB as shown in Fig. \ref{fig:affb} (b). The SFFB is an activation-free module that efficiently captures the periodic structural features of spacecraft noise images by leveraging the Fast Fourier Transform (FFT). Furthermore, SFFB can also extract global features of spacecraft image. Next, we will introduce the detailed process of SFFB.

Firstly, we utilize 2-D Fast Fourier Transform (FFT) to transform the spatial features into frequency domain.
\begin{eqnarray}
    X_{FFT} = FFT2D(X_{SG_1})
\end{eqnarray}
where $X_{SG_1}$ is the input spatial feature maps from the Simple Gate layer in front, $X_{FFT}$ is the frequency features, and $FFT2D($$\cdot$$)$ denotes the 2-D FFT. 

Secondly, in order to make our block more efficient, we no longer use activation functions, but only retain a convolution layer.
\begin{eqnarray}
    X_{C} = Conv(X_{FFT})
\end{eqnarray}
where $X_{C}$ is the frequency feature maps after a $1 \times 1$ frequency convolution layer $Conv($$\cdot$$)$.

Finally, we utilize inverse 2-D FFT to transform frequency domain features back into spatial domain, and insert a skip connection.
\begin{eqnarray}
    X_{SFFB} = InvFFT2D(X_{C}) + X_{SG_1}
\end{eqnarray}
where $X_{SFFB}$ is the output spatial feature maps after inverse 2-D FFT and skip connection, and $InvFFT2D($$\cdot$$)$ denotes the inverse 2-D FFT.

\subsubsection{Simple Gate}
To reduce the computational complexity and improve the efficiency of our model, we utilize Simple Gate to replace the nonlinear activation functions in the model. As shown in Fig. \ref{fig:sg}, in Simple Gate, the feature map is directly divided into two parts in the channel dimension and then multiplied together. Formally, the process is represented as:
\begin{eqnarray}
    SG(X_{in}) = X_1 \odot X_2
\end{eqnarray}
where $X_1$ and $X_2$ are the two equal parts of the input feature map that are divided, $\odot$ denotes the Hadamard product.

\begin{table*}[!ht]
\caption{Quantitative results on LIGHTBOX dataset. The best and the second best results are marked in \textcolor{red}{red} and \textcolor{blue}{blue}.}
\label{tab:lightbox_results}
\centering
\begin{tabular}{lllllll}
\hline
\multirow{2}{*}{Methods}& \multicolumn{2}{c}{$\sigma$=50}  & \multicolumn{2}{c}{$\sigma$=75} & \multicolumn{2}{c}{$\sigma$=100}\\ 
  & \multicolumn{1}{c}{PSNR$\uparrow$}  & \multicolumn{1}{c}{SSIM$\uparrow$} & \multicolumn{1}{c}{PSNR$\uparrow$}  & \multicolumn{1}{c}{SSIM$\uparrow$} & \multicolumn{1}{c}{PSNR$\uparrow$}  & \multicolumn{1}{c}{SSIM$\uparrow$}\\
\hline
DnCNN~\cite{zhang2017beyond} & 25.61 & 0.5685 & 27.72 & 0.5719 & 24.86 & 0.5456 \\
DRUNet~\cite{zhang2021plug} & 30.46 & 0.6078 & 29.98 & 0.5915 & 29.54 & 0.5810 \\
MWDCNN~\cite{tian2023multi} & 30.38 & 0.6070 & 29.84 & 0.5901 & 29.38 & 0.5798 \\
ADNet~\cite{tian2020attention} & 28.87 & 0.5359 & 29.13 & 0.5610 & 27.70 & 0.5479 \\
DRANet~\cite{wu2024dual} & 30.49 & 0.6083 & 29.99 & 0.5914 & 29.62 & 0.5818 \\
SwinIR~\cite{liang2021swinir} & 30.51 & \textcolor{blue}{0.6087} & 30.03 & 0.5924 & 29.63 & 0.5825 \\
SERT~\cite{li2023spectral} & 30.50 & 0.6086 & 30.03 & 0.5924 & 29.66 & \textcolor{blue}{0.5829} \\
EWT~\cite{li2024ewt} & \textcolor{blue}{30.52} & \textcolor{blue}{0.6087} & \textcolor{blue}{30.06} & \textcolor{blue}{0.5927} & \textcolor{blue}{29.69} & \textcolor{blue}{0.5829}\\
RAMiT~\cite{choi2024reciprocal} & 30.39 & 0.6053 & 29.98 & 0.5917 & 29.60 & 0.5819\\
SAFFN(Our approach) & \textcolor{red}{30.53} & \textcolor{red}{0.6092} & \textcolor{red}{30.09} & \textcolor{red}{0.5932} & \textcolor{red}{29.74} & \textcolor{red}{0.5838} \\
\hline
\end{tabular}
\end{table*}

\begin{figure*}[t]
\centering
\begin{minipage}[t]{0.7\linewidth}
  \centering
  \includegraphics[width=1.0\textwidth]{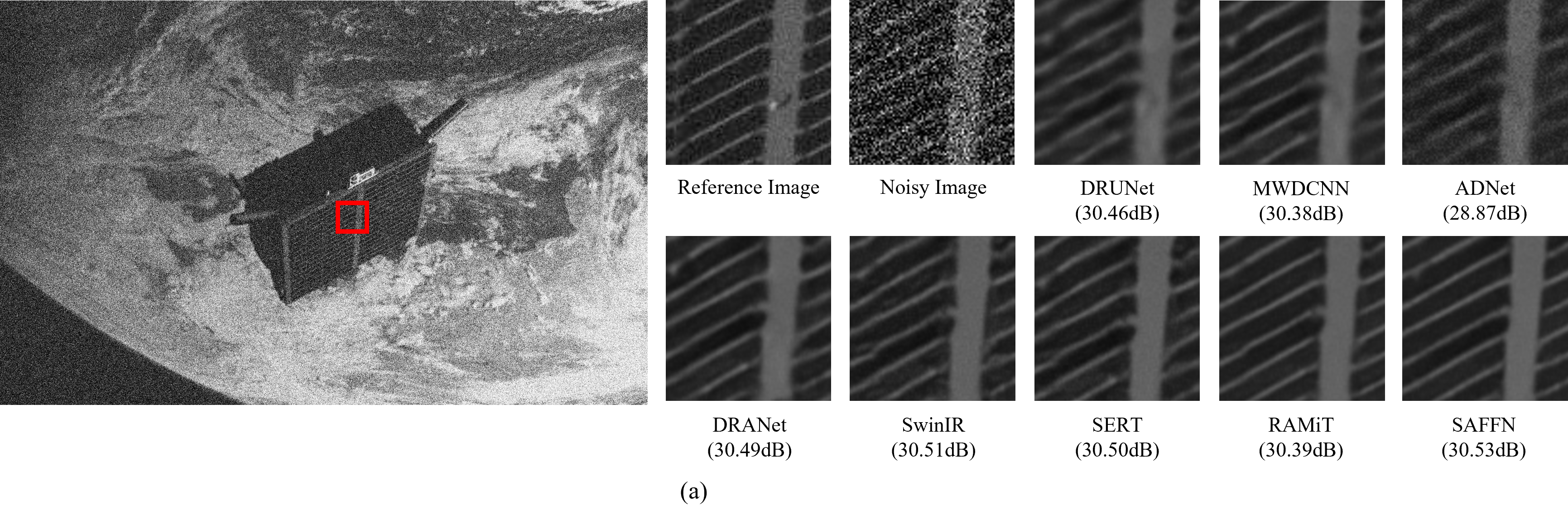}
\end{minipage}  

\begin{minipage}[t]{0.7\linewidth}
  \centering
  \includegraphics[width=1.0\textwidth]{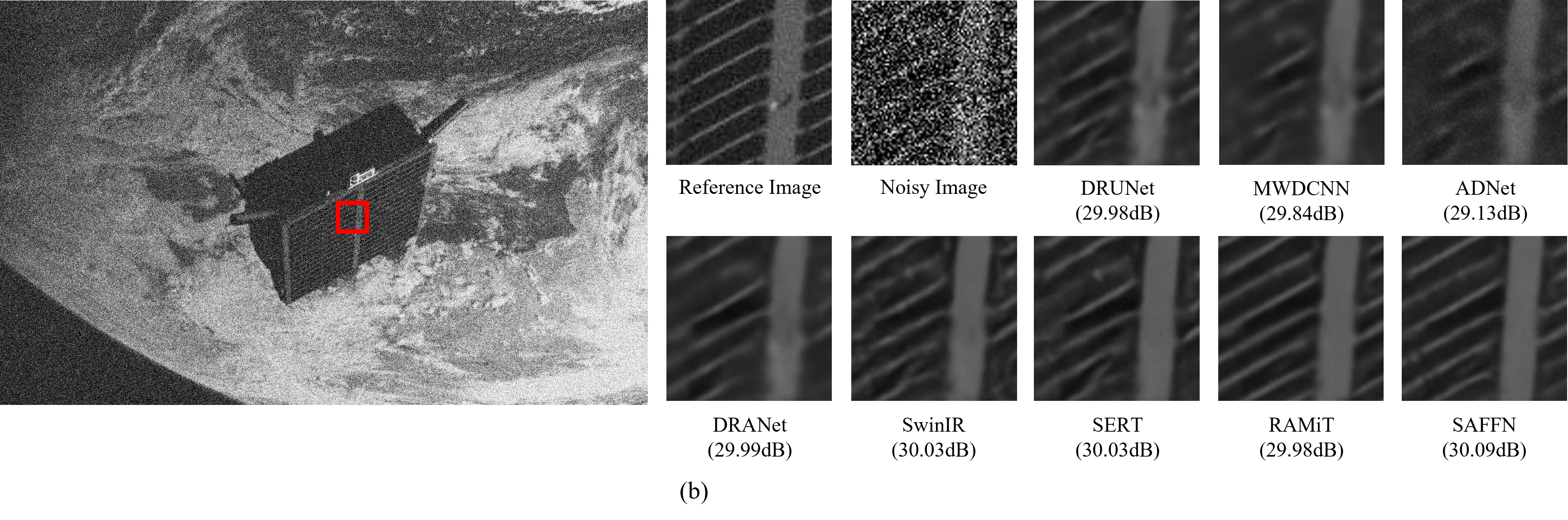}
\end{minipage}  

\begin{minipage}[t]{0.7\linewidth}
  \centering
  \includegraphics[width=1.0\textwidth]{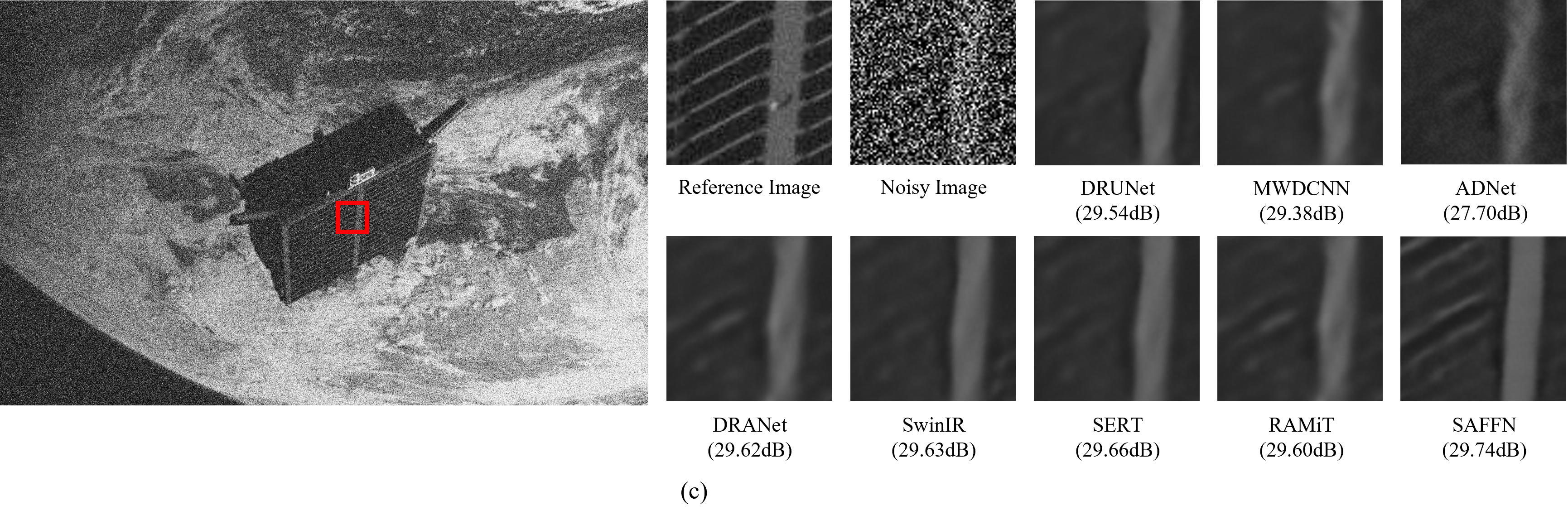}
\end{minipage}  
\caption{Visual comparison on LIGHTBOX. (a): $\sigma$ = 50. (b): $\sigma$ = 75. (c): $\sigma$ = 100.}
\label{fig:visual_results_lightbox}
\end{figure*}

\section{Experiment}
\label{exp}
\subsection{Experimental Settings}
In this section, we will introduce the datasets, evaluation metrics, and implementation details.

\subsubsection{Datasets}
In this paper, we employ the Spacecraft PosE Estimation Dataset (SPEED)~\cite{sharma2020neural} as the training and testing dataset. SPEED is a public PRISMA Tango satellite image dataset, originally used for the spacecraft pose estimation challenge organized by European Space Agency (ESA) and Stanford University's Space Rendezvous Lab (SLAB). It contains two types of data: spacecraft images captured by camera sensors under simulated real space conditions and augmented reality images fused with synthetic images and actual space images. In SPEED, 12000 images are used for training and 2998 images for evaluation, with a resolution of $1920 \times 1200$ pixels for each image. We add Gaussian noise to SPEED images, and build the original-noisy image pairs. The noisy levels include 50, 75 and 100.

In addition, we utilize the LIGHTBOX dataset in the next generation Spacecraft PosE Estimation Dataset (SPEED+)~\cite{park2023satellite} as an additional testing dataset. SPEED+ is also a public spacecraft image dataset and originally used for the second international satellite pose estimation competition organized by ESA and SLAB. The LIGHTBOX dataset in SPEED+ is constructed by simulating real space lighting conditions using a diffuse 
reflection lightbox system, and includes 6740 images, with a resolution of $1920 \times 1200$ pixels for each image. We also add Gaussian noise to LIGHTBOX images, and the noisy levels include 50, 75 and 100.

\subsubsection{Evaluation Metrics}
There are currently many evaluation metrics to evaluate the performance of image denoising~\cite{chen2014quality,zhou2021image}, we adopt two common metrics: Peak Signal-to-Noise Ratio (PSNR) and Structural Similarity (SSIM). The higher values of these two metrics indicate the better performance of the method. Furthermore, we adopt Multiply-Accumulate Operations (MACs) and Runtime to measure the computational complexity and computing efficiency. The lower value of the MACs indicates the lower the computational complexity and the higher the computing efficiency of the method. 

\subsubsection{Implementation Details}
In our proposed SAFFN, the feature channel number is set to 64. The encoder block configuration is [2,2,4,8], the decoder block configuration is [2,2,2,2], and the number of middle block is 12.

For the training, we use NVIDIA GeForce RTX 3090 GPU to train our SAFFN. The total iter number is set to 330000. The initial learning rate is set to 1e-3, and decay in each 1000 iter by TrueCosineAnnealingLR. The optimizer is AdamW($\beta_1$=0.9, $\beta_2$=0.9). The loss function is PSNR loss, and loss weight is 1.0. The batch size is set to 4. The crop size of the GT image is set to 256. 

\begin{table*} [t]
\caption{Effects of proposed modules on SPEED dataset.}
\label{tab:module_ablation}
\centering
\begin{tabular}{lllllll}
\hline
\multirow{2}{*}{Modules}& \multicolumn{2}{c}{$\sigma$=50}  & \multicolumn{2}{c}{$\sigma$=75} & \multicolumn{2}{c}{$\sigma$=100}\\ 
  & \multicolumn{1}{c}{PSNR}  & \multicolumn{1}{c}{SSIM} & \multicolumn{1}{c}{PSNR}  & \multicolumn{1}{c}{SSIM} & \multicolumn{1}{c}{PSNR}  & \multicolumn{1}{c}{SSIM}\\
\hline
Baseline & 31.84 & 0.6912 & 31.26 & 0.6762 & 30.81 & 0.6662 \\
Baseline+SMB  & 31.85 & 0.6915 & 31.29 & 0.6769 & 30.84 & 0.6668 \\
Baseline+SFFB  & 31.85 & 0.6918 & 31.30 & 0.6771 & 30.59 & 0.6669 \\
Baseline+SMB+SFFB & \pmb{31.86} & \pmb{0.6920} & \pmb{31.32} & \pmb{0.6775} & \pmb{30.87} & \pmb{0.6673}\\
\hline
\end{tabular}
\end{table*}

\begin{table*} [t]
\caption{Paired t-test of PSNR performance for the proposed modules ablation experiments.}
\label{tab:statistical_analysis}
\centering
\begin{tabular}{lll}
\hline
  Pairs & t-statistic & p-value\\
\hline
Baseline vs. Baseline+SMB & 57.51 & $<$0.01 \\
Baseline vs. Baseline+SFFB & 66.19 & $<$0.01 \\
Baseline+SMB vs. Baseline+SMB+SFFB  & 71.07 & $<$0.01 \\
Baseline+SFFB vs. Baseline+SMB+SFFB  & 63.72 & $<$0.01 \\
Baseline vs. Baseline+SMB+SFFB  & 76.11 & $<$0.01\\
\hline
\end{tabular}
\end{table*}

\begin{figure}[t]
\centering

\begin{minipage}[t]{1.0\linewidth}
  \centering
  \includegraphics[width=1.0\textwidth]{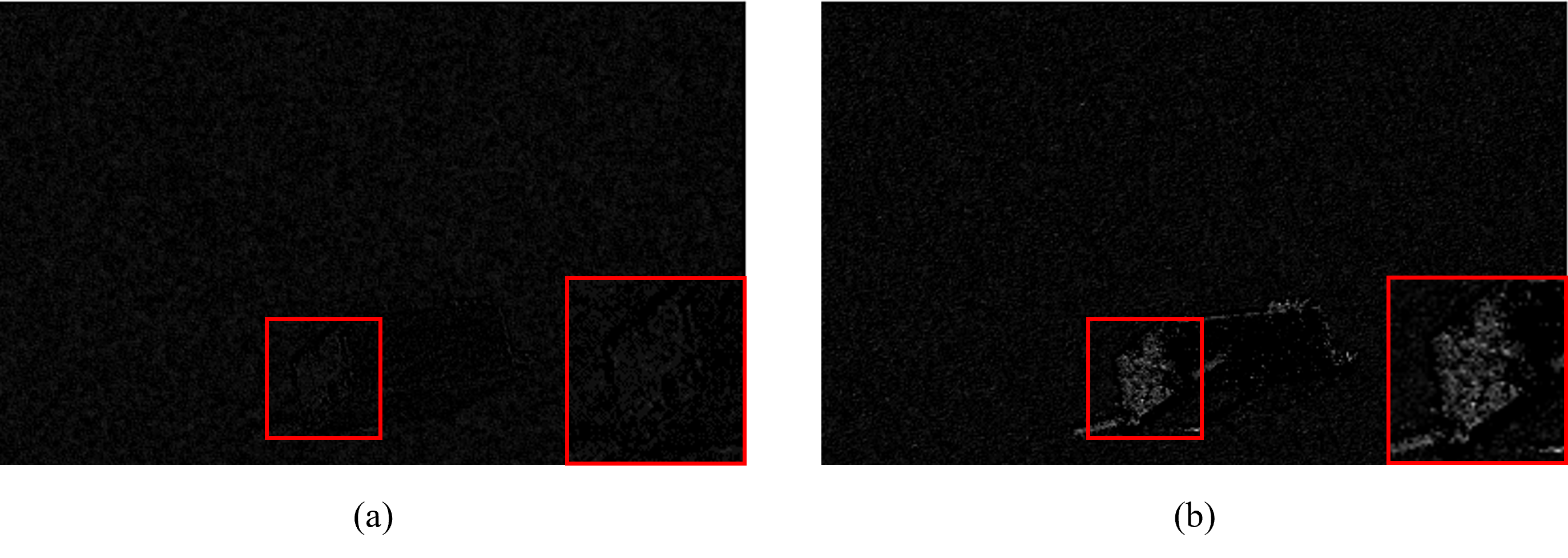}
\end{minipage}  

\caption{Visualization of feature maps. (a): Baseline. (b): Baseline+SMB.}
\label{fig:ablation_intermediate_example1}
\end{figure}

\begin{figure}[t]
\centering

\begin{minipage}[t]{1.0\linewidth}
  \centering
  \includegraphics[width=1.0\textwidth]{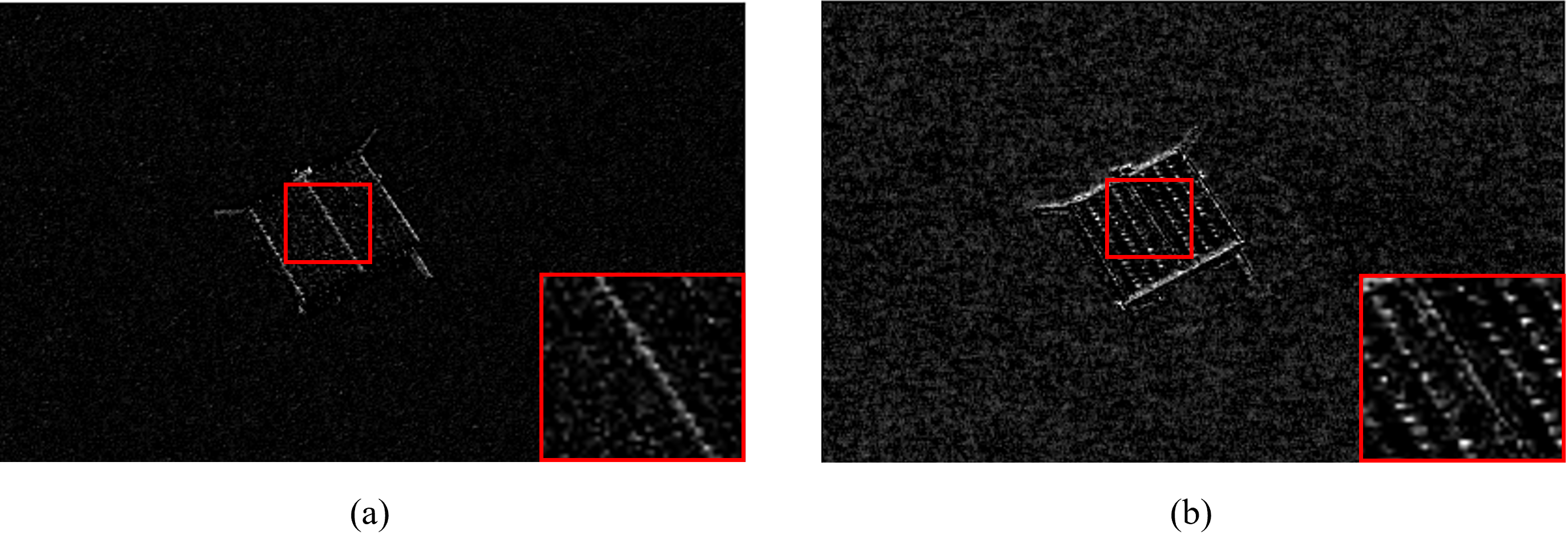}
\end{minipage}  

\caption{Visualization of feature maps. (a): Baseline+SMB. (b): Baseline+SMB+SFFB.}
\label{fig:ablation_intermediate_example2}
\end{figure}

\subsection{Comparison Results of the SPEED dataset}
\label{sec: speed_results}
In this section, we implement several common image denoising methods and retrain them on the SPEED dataset to ensure the fair and valid comparisons. Three CNN-based methods (DnCNN~\cite{zhang2017beyond}, DRUNet~\cite{zhang2021plug}, MWDCNN~\cite{tian2023multi}), two attention-based methods(ADNet~\cite{tian2020attention}, DRANet~\cite{wu2024dual}) four Transformer-based methods (SwinIR~\cite{liang2021swinir}, SERT~\cite{li2023spectral}, EWT~\cite{li2024ewt}, RAMiT~\cite{choi2024reciprocal}) and compared them with our proposed SAFFN. Table \ref{tab:speed_results} presents the quantitative comparison results with three different noisy levels ($\sigma$ = 50, 75, 100). We can evidently observe that our SAFFN significantly outperforms the other image denoising methods on the SPEED test dataset, demonstrating the superiority of our approach. Specifically, compared to the current state-of-the-art method (SwinIR), our SAFFN achieves performance gains of 0.03dB, 0.07dB, 0.09dB in PSNR and 0.0011, 0.0015, 0.0016 in SSIM on three different noisy levels ($\sigma$ = 50, 75, 100). Compared to DRUNet, our SAFFN achieves performance gains of 0.07dB, 0.10dB, 0.15dB in PSNR and 0.0016, 0.0018, 0.0026 in SSIM. In addition to the performance improvement, we can also observe a significant reduction in the computational complexity of SAFFN. Specifically, our SAFFN achieves a reduction in MACs by approximately 2.5 times compared to the second best method SwinIR and achieves a reduction in Runtime by astonishing 33.7 times compared to SwinIR.

\subsubsection{Visual Results of the SPEED dataset}
To further demonstrate the effectiveness of our SAFFN, we present the spacecraft image denoising visual results on the SPEED dataset in Fig. \ref{fig:visual_results_img1} and Fig. \ref{fig:visual_results_img2}. The patches for highlighting are marked with red boxes. From the visual results in Fig. \ref{fig:visual_results_img1}, we can observe that the structure of edge components in the images denoised by other methods are still relatively blurry. From the visual results in Fig. \ref{fig:visual_results_img2},  we can observe that the repetitive periodic structure such as cells of spacecraft solar panel in the images denoised by other methods are still unclear and incomplete. In contrast, the aforementioned structures are more clear and complete in the images denoised using our SAFFN, which indicates that our SAFFN can effectively model the edge structures and the repetitive periodic features in spacecraft noise image.

\subsection{Comparison Results of the LIGHTBOX dataset}
In this section, we evaluate the performance of our proposed SAFFN on the LIGHTBOX dataset (the SAFFN model trained on the SPEED dataset is directly tested without further fine-tuning) and compare with the nine methods mentioned in Section \ref{sec: speed_results}. The quantitative comparison results with three different noisy levels ($\sigma$ = 50, 75, 100) are shown in Table \ref{tab:lightbox_results}. We can observe that our SAFFN still significantly outperforms the other image denoising methods on the LIGHTBOX dataset, demonstrate the effectiveness and stronger generalization of our SAFFN.

\subsubsection{Visual Results of the LIGHTBOX dataset}
We also present the spacecraft image denoising visual results on the LIGHTBOX dataset in Fig. \ref{fig:visual_results_lightbox}. From the visual results, we can clearly observe that the structural details of the spacecraft in the image after denoising with our SAFFN are significantly clearer. Particularly for the denoising results of images with the noisy level of 100, other image denoising methods are unable to restore the basic contour of the spacecraft, while our SAFFN successfully recovers the approximate structure of spacecraft solar panel.

\begin{table} [t]
\caption{Effects of different edge convolution kernel design methods on SPEED dataset.}
\label{tab:kernel_ablation1}
\centering
\begin{tabular}{lll}
\hline
Modules & PSNR & SSIM \\
\hline
2-kinds & 31.30 & 0.6773 \\
4-kinds & 31.32 & 0.6775  \\
8-kinds & \pmb{31.33} & \pmb{0.6776} \\
\hline
\end{tabular}
\end{table}

\begin{table} [t]
\caption{Effects of different edge convolution kernel design methods on LIGHTBOX dataset.}
\label{tab:kernel_ablation2}
\centering
\begin{tabular}{lll}
\hline
Modules & PSNR & SSIM \\
\hline
2-kinds & 30.08 & 0.5931 \\
4-kinds & \pmb{30.09} & \pmb{0.5932} \\
8-kinds & 30.08 & 0.5930 \\
\hline
\end{tabular}
\end{table}

\begin{table} [t]
\caption{Performance comparison between SFFB and CFFB on SPEED dataset.}
\label{tab:sffb_ablation2}
\centering
\begin{tabular}{llll}
\hline
Modules & PSNR & SSIM & MACs \\
\hline
SFFB & \pmb{31.32} & \pmb{0.6775} & \pmb{86.36G} \\
CFFB & 30.55 & 0.6764 & 107.10G \\
\hline
\end{tabular}
\end{table}

\subsection{Ablation Studies}
In this section, we conduct the ablation studies of proposed modules, different edge convolution kernel design methods and SFFB design rationality.

\subsubsection{Ablation Study of proposed modules}
We present the ablation experiments on the SPEED dataset to verify the effectiveness of our proposed modules in SAFFN. The results reported on the SPEED dataset under multiple noisy levels are shown in Table \ref{tab:module_ablation}. The first row describes the Baseline model, which is the common nonlinear activation free network inspired by~\cite{chen2022simple}. In the second row, we add the SMB module, which brings a performance gain as SMB can better model the structure of spacecraft. In the third row, we add the SFFB module, which also brings a performance gain as SFFB can better extract periodic features from spacecraft image. However, SFFB is difficult to model the spacecraft structure under high noise ($\sigma$=100). In the fourth row, we add both SFFB and SMB simultaneously, which further improves the performance, and solves the problem of modeling difficulties in SFFB under high noise.

To statistically verify the significance of these improvements, we conduct the paired t-test on the PSNR performance of proposed modules ablation experiments with noisy level 75. The results are shown in Table \ref{tab:statistical_analysis}. The p-values of all five paired t-tests are less than 0.05, signifying a significant PSNR performance difference among these models. The statistical analysis clearly demonstrates the effectiveness of our proposed modules in SAFFN.

To intuitively demonstrate our proposed modules, we also provide the visualization results of intermediate feature maps for models with different modules in Fig. \ref{fig:ablation_intermediate_example1} and Fig. \ref{fig:ablation_intermediate_example2}. In Fig. \ref{fig:ablation_intermediate_example1}, we can observe that the edge contour features of spacecraft become sharper after adding the SMB module, indicating that our proposed SMB can effectively extract the edge features and model the structures in spacecraft image. In Fig. \ref{fig:ablation_intermediate_example2}, we can observe that the structure of spacecraft solar panel becomes more complete after adding the SFFB module, suggesting that our proposed SFFB can effectively extract the periodic repetitive features in spacecraft image.

\subsubsection{Ablation study of the effect of different edge convolution kernel design methods}
We examine the effect of different edge convolution kernel design methods: 2-kinds (vertical, horizontal), 4-kinds (vertical, horizontal, diagonal with 2 directions) and 8-kinds (vertical, horizontal, diagonal with 4 directions) on the SPEED dataset and LIGHTBOX dataset, the results for noisy level 75 are shown in Table \ref{tab:kernel_ablation1} and Table \ref{tab:kernel_ablation2}. We find that our 4-kinds kernel design outperforms 2-kinds kernel design on both the SPEED dataset and LIGHTBOX dataset. We also find that our 4-kinds kernel design performs slightly worse than 8-kinds kernel design on the SPEED dataset, but outperforms than 8-kinds kernel design on the LIGHTBOX dataset, indicating that our 4-kinds kernel design has stronger generality.

\subsubsection{Ablation study of the rationality of SFFB design}
We examine the rationality of SFFB design by comparing the performance of SFFB and Complex Fast Fourier Block (CFFB, replacing the single $1 \times 1$ convolution layer with two $1 \times 1$ convolution layers and activation functions) on the SPEED dataset, the results for noisy level 75 are shown in Table \ref{tab:sffb_ablation2}. Compared to CFFB, our SFFB achieves performance gains of 0.77dB in PSNR and 0.0011 in SSIM with a reduction in MACs. In other words, our SFFB achieves performance improvement while reducing computational complexity. Therefore, the design of our SFFB is rationality.

\section{Conclusion}
\label{conclusion}
In this paper, we propose a novel and efficient spacecraft image denoising method --- Structure modeling Activation Free Fourier Network (SAFFN), which can better accommodate to the unique characteristics of spacecraft image. In SAFFN, we design an Structure Modeling Block (SMB) to extract the edge features and model the structure, enhancing the ability to distinguish spacecraft components in the low-light spacecraft noise image. Furthermore, we design an Activation Free Fourier Block (AFFB) that utilizes Simplified Fast Fourier Block (SFFB) to obtain the repetitive periodic features in spacecraft image. Extensive ablation experiments demonstrate the effectiveness of our proposed modules. Simultaneously, experimental results on the SPEED dataset and LIGHTBOX dataset substantiate that our SAFFN not only significantly outperforms the existing state-of-the-art image denoising methods, but also achieves lower computational complexity.

This paper also has some limitations. Specifically, we have not yet explored the performance of SAFFN under extreme noise conditions (noisy level $\sigma>100$), nor have we further validated its generalization ability on real-world spacecraft image datasets. Therefore, future work will focus on improving the performance of our SAFFN on real-world spacecraft image datasets (especially those obtained under extreme noise conditions).

\section*{CRediT authorship contribution statement}
\pmb{Jingfan Yang:} Conceptualization, Data curation, Investigation, Methodology, Software, Writing – original draft. \pmb{Hu Gao:} Conceptualization, Formal analysis, Writing – review \& editing. \pmb{Ying Zhang:} Investigation, Validation, Conceptualization. \pmb{Bowen Ma:} Validation, Formal analysis, Writing – review \& editing. \pmb{Depeng Dang:} Funding acquisition, Supervision, Validation, Writing – review \& editing.

\section*{Declaration of competing interest}
The authors declare that they have no known competing financial interests or personal relationships that could have appeared to influence the work reported in this paper.

\section*{Data availability}
Data will be made available on request.

\section*{Acknowledgement}
This work was supported by the National Key Research and Development Program of China (Grant 2020YFC1523303).




\bibliographystyle{elsarticle-num} 
\bibliography{saffn}

\subsection*{  } 
\setlength\intextsep{0pt} 
\begin{wrapfigure}{l}{25mm}
    \centering
    \includegraphics[width=1in,height=1.25in,clip,keepaspectratio]{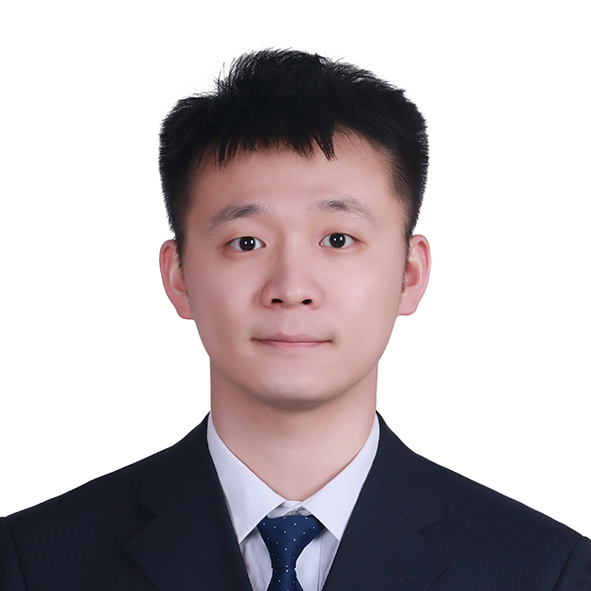}
\end{wrapfigure}
\noindent \textbf{Jingfan Yang} is currently pursuing the Ph.D. degree
with the School of Artificial Intelligence, Beijing
Normal University, Beijing, China. His research interests include image restoration and image enhancement.\par

\hspace*{\fill} 

\subsection*{  } 
\setlength\intextsep{0pt} 
\begin{wrapfigure}{l}{25mm}
    \centering
    \includegraphics[width=1in,height=1.25in,clip,keepaspectratio]{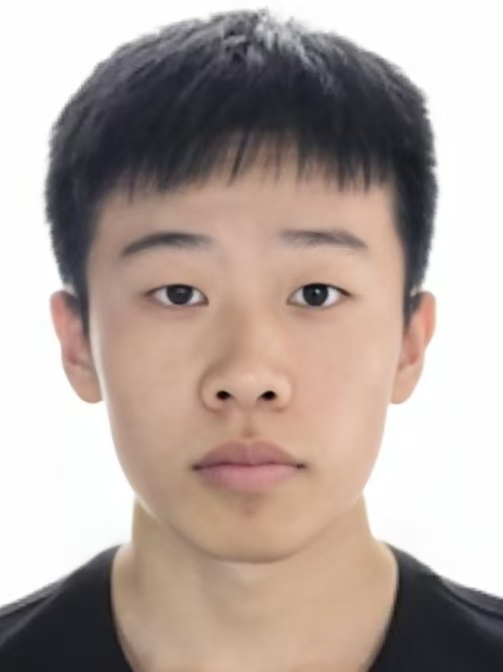}
\end{wrapfigure}
\noindent \textbf{Hu Gao} is currently pursuing the Ph.D. degree
with the School of Artificial Intelligence, Beijing
Normal University, Beijing, China. His research interests include image restoration and image enhancement.\par

\hspace*{\fill} 

\subsection*{  } 
\setlength\intextsep{0pt} 
\begin{wrapfigure}{l}{25mm}
    \centering
    \includegraphics[width=1in,height=1.25in,clip,keepaspectratio]{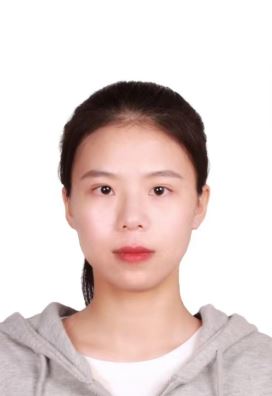}
\end{wrapfigure}
\noindent \textbf{Ying Zhang} is currently pursuing the Ph.D. degree with the School of Artificial Intelligence, Beijing Normal University, Beijing, China. Her research interests include relation extraction and multi modal fusion.\par

\hspace*{\fill} 

\subsection*{  } 
\setlength\intextsep{0pt} 
\begin{wrapfigure}{l}{25mm}
    \centering
    \includegraphics[width=1in,height=1.25in,clip,keepaspectratio]{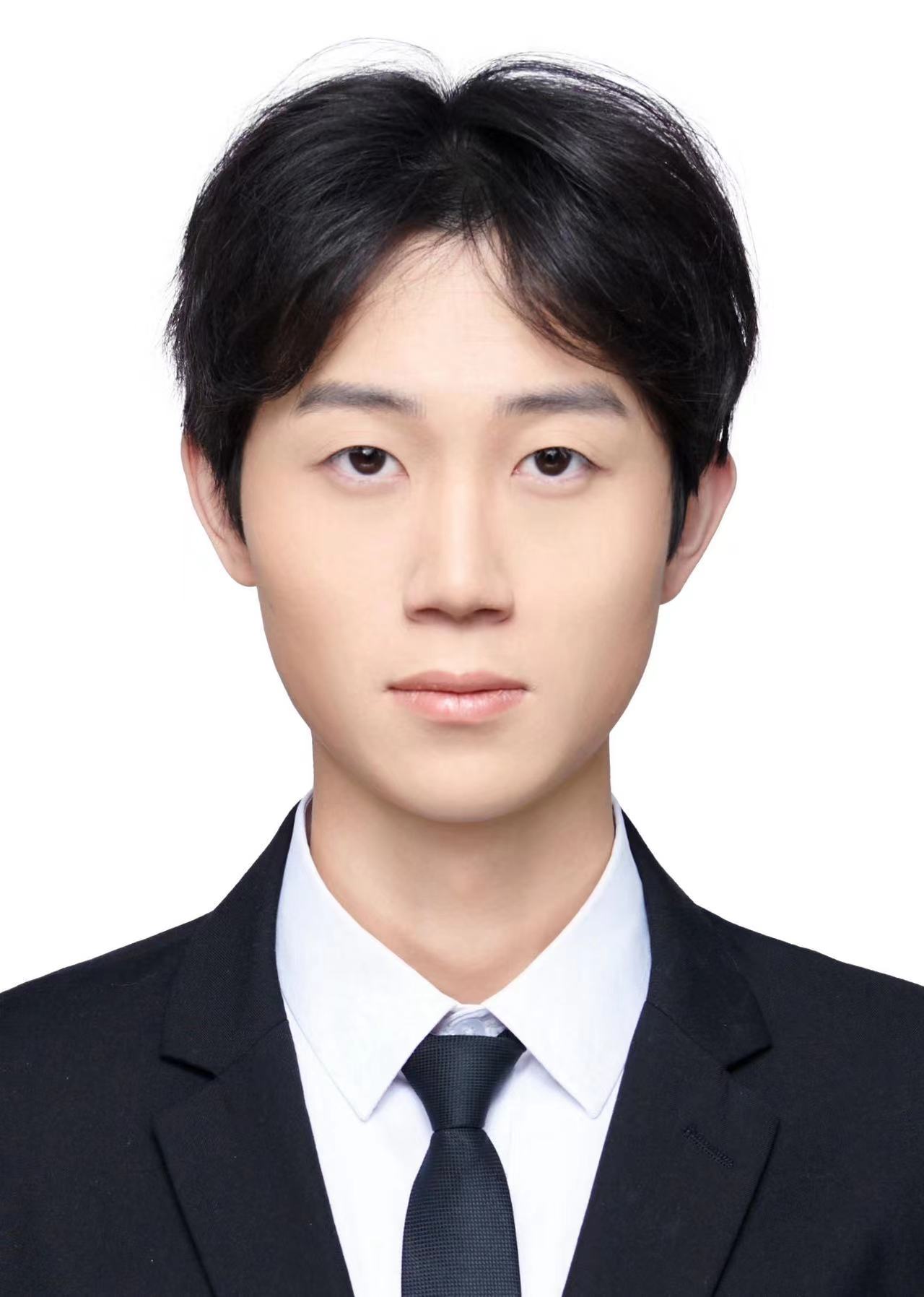}
\end{wrapfigure}
\noindent \textbf{Bowen Ma} is currently pursuing the Master degree with the School of Artificial Intelligence, Beijing Normal University, Beijing, China. His research interests include multi modal fusion and image enhancement.\par

\hspace*{\fill} 

\subsection*{  } 
\setlength\intextsep{0pt} 
\begin{wrapfigure}{l}{25mm}
    \centering
    \includegraphics[width=1in,height=1.25in,clip,keepaspectratio]{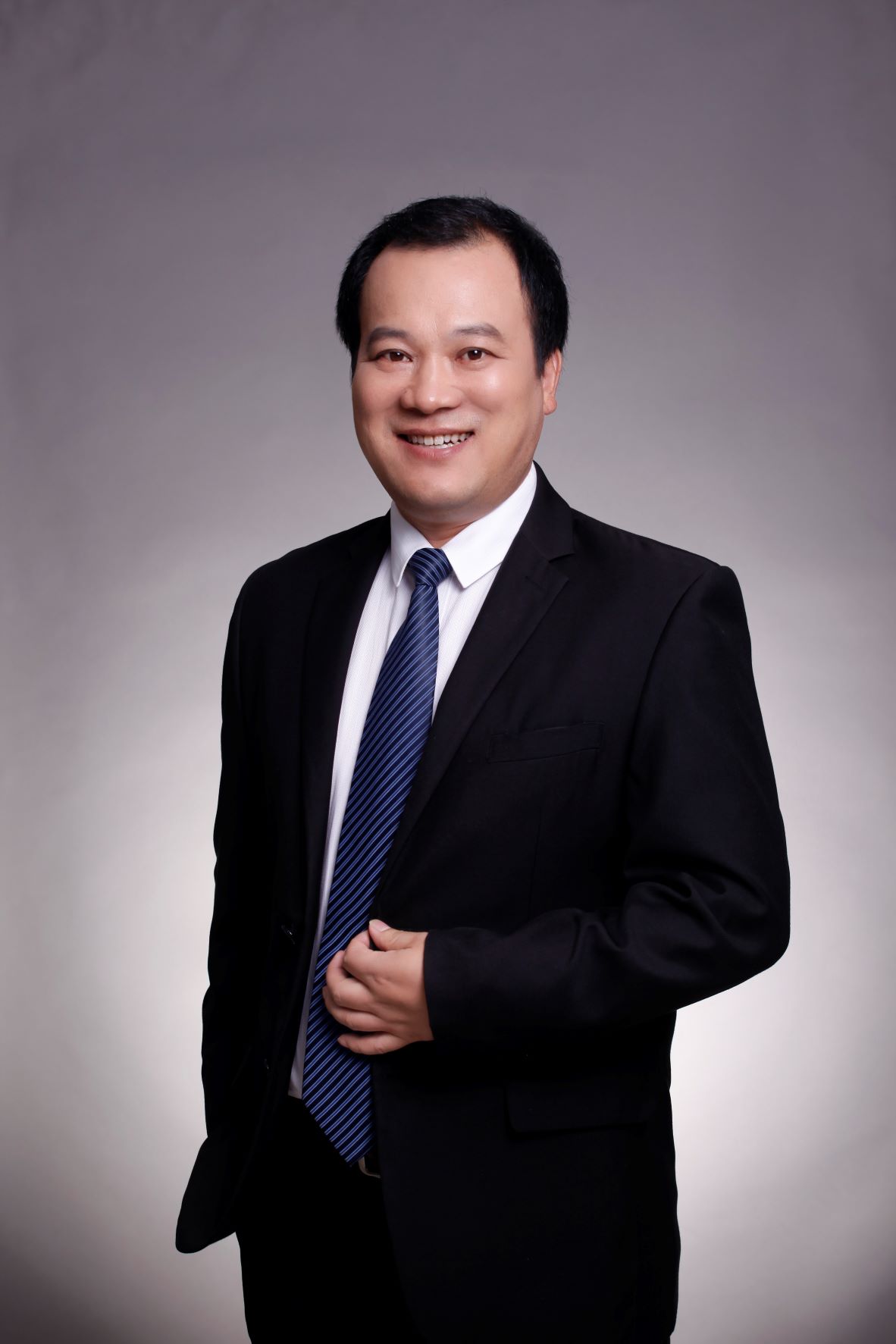}
\end{wrapfigure}
\noindent \textbf{Depeng Dang} receive the Ph.D degree in computer science and technology from the Huazhong University of Science and Technology, Wuhan, China, in 2003. From July 2003 to June 2005, he did his postdoctoral research with the Department of Computer Science and Technology, Tsinghua University, China. Currently, he is a full professor and supervisor of PhD students in computer science from Beijing Normal University, China.\par
  
\end{document}